\documentclass{article} % For LaTeX2e
\usepackage{iclr2026_conference,times}
\usepackage{graphicx}
\usepackage{subcaption}
\usepackage{multirow}
\usepackage{amsmath,amsfonts,bm}

\def\eqref#1{equation~\ref{#1}}
\def\1{\bm{1}}

\DeclareMathAlphabet{\mathsfit}{\encodingdefault}{\sfdefault}{m}{sl}
\SetMathAlphabet{\mathsfit}{bold}{\encodingdefault}{\sfdefault}{bx}{n}

\usepackage{hyperref}
\usepackage{url}

\title{
OmniPatch: A Universal Adversarial Patch 
for ViT-CNN Cross-Architecture Transfer in Semantic Segmentation
}

\author{
Aarush Aggarwal\thanks{These authors contributed equally to this work.}, Akshat Tomar\footnotemark[1], Amritanshu Tiwari\thanks{These authors also contributed equally to this work.}, Sargam Goyal\footnotemark[2] \\
Data Science Group, Indian Institute of Technology Roorkee, India \\
\texttt{\{aarush\_a@ma, akshat\_t@mfs, amritanshu\_t@mfs, sargam\_g@mfs\}.iitr.ac.in}
}
\iclrfinalcopy % Uncomment for camera-ready version, but NOT for submission.
\begin{document}
\maketitle
\begin{abstract}
Robust semantic segmentation is crucial for safe autonomous driving, yet deployed models remain vulnerable to black-box adversarial attacks when target weights are unknown. Most existing approaches either craft image-wide perturbations or optimize patches for a single architecture, which limits their practicality and transferability. We introduce \textbf{OmniPatch}, a training framework for learning a \emph{universal adversarial patch} that generalizes across images and both ViT and CNN architectures without requiring access to target model parameters.
\end{abstract}
\section{Introduction \& Related Work}
Deep learning has transformed computer vision, enabling significant advancements in image classification \citep{NIPS2012_c399862d}, object detection \citep{Girshick_2014_CVPR}, and semantic segmentation \citep{long2015fully} tasks. In autonomous driving, segmentation models provide essential pixel-level understanding for planning and control. However, these networks remain vulnerable to adversarial examples that induce catastrophic failures \citep{szegedy2013intriguing, goodfellow2014explaining}. % references related to NN being vulnerable to adversarial examples
Consequently, a thorough analysis of these vulnerabilities is imperative for creating of robust, safety-critical systems.
% Broad overview and listing that didn't find any work that transfers patch from ViT to CNN

% 1. Patch instead of whole image perturbation\\
% Patch deployment makes more sense in real life scenarios
% \\
% 2. Transferability of patches\\
% Transferability across models, across different architectures, and across images with EoT.\\

% Para-1 for in general semantic segmentaion, adversarial attakcs

% Despite recent progress in adversarial attacks on segmentation models %citatations here should include general attacks older 
% ~\citep{DBLP:journals/corr/abs-2111-11368,jia2023transegpgdimprovingtransferabilityadversarial,he2024transferable}, but two key limitations remain. Most methods generate image-wide perturbations, which are hard to deploy physically and require training multiple examples across scenes, and they typically optimize for a single model, offering limited insight into cross-architecture transfer.

% Para-2 Attacks with image-wide perturbation and list very few patch attacks exist
Since the foundational demonstration of adversarial vulnerabilities in segmentation models \citep{xie2017adversarial,hendrik2017universal}, the field has matured significantly, yet a critical limitation persists regarding actual deployability of these attacks. The majority of existing methods, including recent approaches \citep{gu2022segpgd, jia2023transegpgd}, rely on image-wide perturbations that are impractical for physical use. Conversely, research on applicable adversarial patches for semantic segmentation remains scarce \citep{rossolini2023real, nesti2022evaluating}, leaving deployable threats largely unexplored.

% Para-3 focuses on transferability of these attacks acorss architectures
Model agnostic adversarial transferability has received limited attention in the context of semantic segmentation \citep{he2024transferable}. %mention if any such papers exist
While CNNs benefit from local biases \citep{geirhos2018imagenet}, ViTs are significantly more vulnerable to patch-based attacks due to their global attention mechanisms. % cite all the papers related to ViT being more vulnerable to attacks
\citep{Mahmood2021Robust,Mo2022AdvTraining,PatchFool2025}. This sensitivity makes ViTs effective training surrogates. The realization of a physically deployable adversarial perturbation capable of operating on heterogeneous segmentation models is a formidable pursuit. To the best of our knowledge, \citet{shekhar2025cross} represents the sole endeavor to address this. However, their marginal generalization leaves reliable universal patch transfer across segmentation architectures unsolved.

% Adversarial transferability in semantic segmentation is especially underexplored across fundamentally different architectures. CNNs, with strong locality and edge-based inductive biases, are often harder to break, whereas ViTs have been shown to be more vulnerable due to their global attention mechanisms~\citep{Mahmood2021Robust,Mo2022AdvTraining,PatchFool2025,Islam2025Mechanistic}. However, most prior work still focuses on classification or within-architecture transfer, leaving the ViT$\rightarrow$CNN (and CNN$\rightarrow$ViT) case in segmentation far less studied.

% Para-4 a bried introductioni to our approach 
We address this gap with \textbf{OmniPatch}, a patch-based adversarial perturbation designed with physical constraints and cross-architectural transferability. We first exploit the ease of destabilizing ViTs, then guide transfer to CNNs through ensemble training with gradient alignment. Using the ViT surrogate, we bias the placement of the patch towards regions of higher uncertainty.

% Para-2 and Para-3 should include all the relevant citations

\section{Method}
\begin{figure}
    \centering
    \includegraphics[width=1\linewidth]{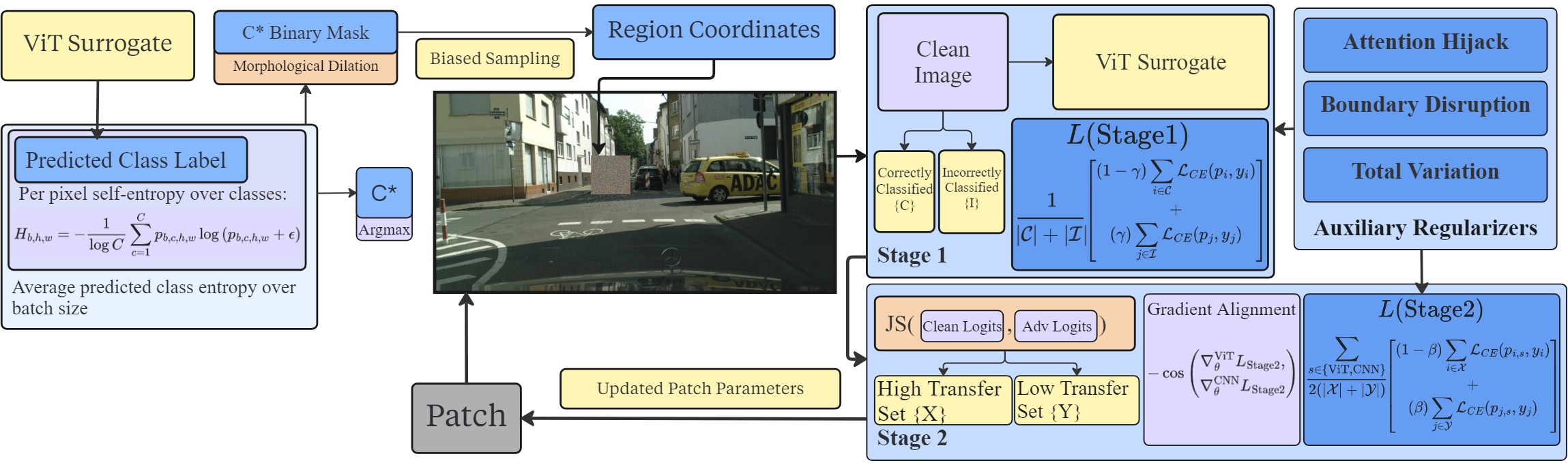}
    \caption{We apply the patch on the most sensitive class and use a two-stage regularized loss, weighting confident ViT pixels, then ViT-CNN ensembling via JS divergence and gradient alignment.}
    \label{fig:method_pipeline}
\end{figure}

% \label{gen_inst}
We formalize a patched adversarial strategy (Figure \ref{fig:method_pipeline}) which uses sensitive region placement with a two-staged objective using ViT and CNN surrogates. We add training regularizers for stability.
\subsection{Sensitive Region Placement}     
\label{srp}
Using a ViT surrogate, we compute class-wise predictive self-entropy on clean images and select the class $c^\star$ with highest uncertainty. We extract the predicted masks for $c^\star$ and apply the morphological dilation by $k$ pixels to expand the feasible placement region. A suitable patch location is then sampled with entropy-biased sampling, where candidate patch locations are weighted by their pixel-wise uncertainty, restricted to the top-$p\%$ . If no valid region for $c^\star$ exists, we sample randomly.\footnote{All methods are mathematically detailed in Appendix \ref{app:method_1}} This positions our patch to exploit the inductive bias gap between the global attention mechanisms of ViT's and the local feature extraction of CNN's.

\subsection{Two Stage Training}
We modify a two-stage training paradigm which was first formulated in \citet{jia2023transegpgd}. We further introduce auxiliary and gradient alignment objectives.

\textit{Stage 1 (ViT-only)}: 
\label{stage1}
We first optimize OmniPatch to destabilize the ViT surrogate, exploiting its sensitivity by targeting its high-confidence predictions.  We use a ($\gamma$) weighted cross-entropy $(\mathcal{L}_{CE})$ objective between prediction ($p$) and label ($y$) that prioritizes clean image pixels ($\mathcal{C}$) which are classified correctly by the model, while down-weighting those already misclassified ($\mathcal{I}$). This loss function guides OmniPatch to induce errors in otherwise confident ViT segmentation outputs:
\begin{equation}
L_{\text{Stage1}} =\frac{1}{|\mathcal{C}| + |\mathcal{I}|}\left[
(1-\gamma){\sum_{i\in\mathcal{C}}\mathcal{L}_{CE}(p_i,y_i)} + (\gamma){\sum_{j\in\mathcal{I}}\mathcal{L}_{CE}(p_j,y_j)}\right],
\end{equation}

\textit{Stage 2 (ViT+CNN ensemble)}:
\label{stage2}
After completion of Stage 1, the second stage extends training to a heterogeneous ViT and CNN ensemble. We define a high transfer set $\mathcal{X}$ containing pixels that exhibit a significant distributional shift (quantified by the Jensen-Shannon divergence between clean and adversarial logits), while the remaining pixels form the low-transfer set $\mathcal{Y}$. The objective is computed as the mean over both surrogates. We ($\beta$) weight these high-divergence pixels ($\mathcal{X}$) relative to the low-divergence ones ($\mathcal{Y}$) to maximize cross-architecture transferability:
\begin{equation}
    L_{\text{Stage2}} = \frac{1}{2 (|\mathcal{X}| + |\mathcal{Y}|)} \sum_{s \in \{\text{ViT}, \text{CNN}\}} \left[ (1-\beta) \sum_{i \in \mathcal{X}} \mathcal{L}_{{CE}}(p_{i,s}, y_i) + (\beta) \sum_{j \in \mathcal{Y}} \mathcal{L}_{{CE}}(p_{j,s}, y_j) \right],
\end{equation}

% Tell the intuition that it helps in making damage, and why are we preferring ViT for stage1 and not cnn\\

% b.	First describe L,H sets are calculated mention why do we changed to js from kl as used in TransegPGD and mention that results are in ablation section.\\
% c.	Any intuition why this majorly work for transferability.\\
% \\

% Novelty for cross architectural transferability: gradient alignment loss (make this look much important)\\
However, standard ensemble training over fundamentally different segmentation models causes gradient updates to interfere destructively. To resolve this destructive interference, we penalize incongruent update directions by maximizing cosine similarity between the gradients of the ViT ($\nabla_{\theta}^{\text{ViT}}L_{\text{Stage2}}$) and the CNN ($\nabla_{\theta}^{\text{CNN}}L_{\text{Stage2}}$)\footnote{Gradients are computed with respect to the patch while keeping surrogate weights frozen} surrogates. This constraint homogenizes the update vectors to prevent conflicting gradient directions:
\begin{equation}
L_{\text{align}} = -\cos\!\big(\nabla_{\theta}^{\text{ViT}}L_{\text{Stage2}}, \nabla_{\theta}^{\text{CNN}}L_{\text{Stage2}}),
\end{equation}
% \\NOTE THAT GRADIENT ALIGNMENT IS ONLY USED IN STAGE-2 Mention this

% To bridge the inductive-bias gap between global (ViT) and local (CNN) representations, we also add a gradient-alignment term with cosine similarity that encourages the patch to follow a common update direction across models:

\subsection{Auxiliary losses/ regularizers}
We introduce three auxiliary objectives to our existing loss function: a) attention hijacking, b) boundary disruption, and c) total variation. \\
\textbf{Attention Hijacking:} We apply the self-attention attack strategy proposed by \citet{naseer2021improving}, which forces the ViT to prioritize the patch over the true label in its internal representation.\\
\textbf{Boundary Disruption:} We invert the boundary loss constraint proposed by \citet{kervadec2019boundary} to induce fragmentation in the segmentation boundaries. \\
\textbf{Total Variation:} We use the anisotropic version of this loss to act as a visual noise control regularizer used in prior work \citep{johnson2016perceptual}

\subsection{Final Objective.}
We use a sequential training schedule by first optimizing $L_{\text{Stage1}}$ for $N_1$ epochs, then switching to $L_{\text{Stage2}}$ for the remaining $N_2$ epochs. Subsidary losses ($L_{\text{attn}}$, $L_{\text{boundary}}$, $L_{\text{TV}}$) are applied throughout the process. The gradient alignment term is activated exclusively during Stage 2 to enforce consistency between the surrogate gradients. The unified objective is defined as:
\begin{equation}
L_{\text{total}} = L_{\text{attack}} + \lambda_{\text{attn}} L_{\text{attn}}  + \lambda_{\text{boundary}} L_{\text{boundary}}+ \lambda_{\text{TV}} L_{\text{TV}} + \displaystyle \1_\mathrm{(\text{Stage2})} \cdot \lambda_{\text{align}} L_{\text{align}},
\end{equation}
where $L_{\text{attack}} \in \{L_{\text{Stage1}}, L_{\text{Stage2}}\}, \displaystyle \1_\mathrm{(.)}$ denotes the indicator function. \\
% 5.	For Robustness across image conditions, we also apply EOT (what in EOT)\\
We also apply Expectation-over-Transformation (EOT) at every step, modeling random scale, rotation, and translation to simulate variations in literal conditions.

\section{Experiments}
Experiments are performed using images from the Cityscapes dataset \citep{cordts2016cityscapes}. \\The Cityscapes dataset consists of street-scene images with pixel-level labels. It has 19 classes with 2,975 training images, 500 validation images, and 1,525 testing images $(2048\times1024)$. As source models, we use PIDNet-S \citep{xu2023pidnet} as the CNN surrogate and SegFormer \citep{xie2021SegFormer} as the ViT surrogate. We use PIDNet-M, PIDNet-L, BiSeNetV1 \citep{yu2018bisenet}, BiSeNetV2 \citep{yu2021bisenet} as target models for performance evaluation and testing.

We train for 20 epochs, split evenly between Stage 1 and Stage 2. Specifically, each epoch contains 300 randomly sampled images with 7 attack iterations. Unless otherwise noted, a $200 \times 200$ ($1.9\%$ area) patch is placed on the sensitive region calculated by ViT surrogate using clean image predictions. The mean Intersection over Union (mIoU) is used to evaluate the adversarial performance. 

\begin{figure*}[h]
\centering
\setlength{\tabcolsep}{2pt}
\renewcommand{\arraystretch}{1}

% ADD THIS LINE - reduces space between image and subcaption
\captionsetup[subfigure]{skip=2pt, font=footnotesize} 
% \centering
% \setlength{\tabcolsep}{2pt}
% \renewcommand{\arraystretch}{0}  % Remove extra row spacing

\begin{minipage}[h]{0.12\linewidth}
    \centering
    \includegraphics[width=\linewidth]{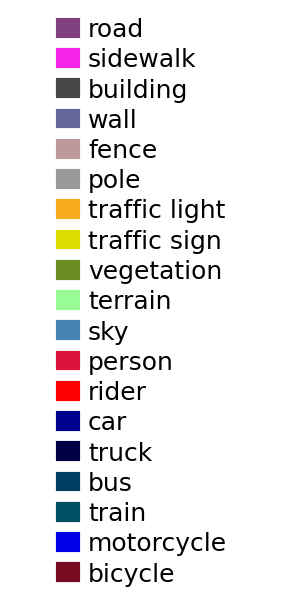}
\end{minipage}\hspace{4pt}%
\begin{minipage}[h]{0.86\linewidth}
\centering
\begin{tabular}{@{}cccc@{}}

% --------- ROW 1 ----------
\begin{minipage}[h]{0.23\linewidth}
    \centering
    \includegraphics[width=\linewidth]{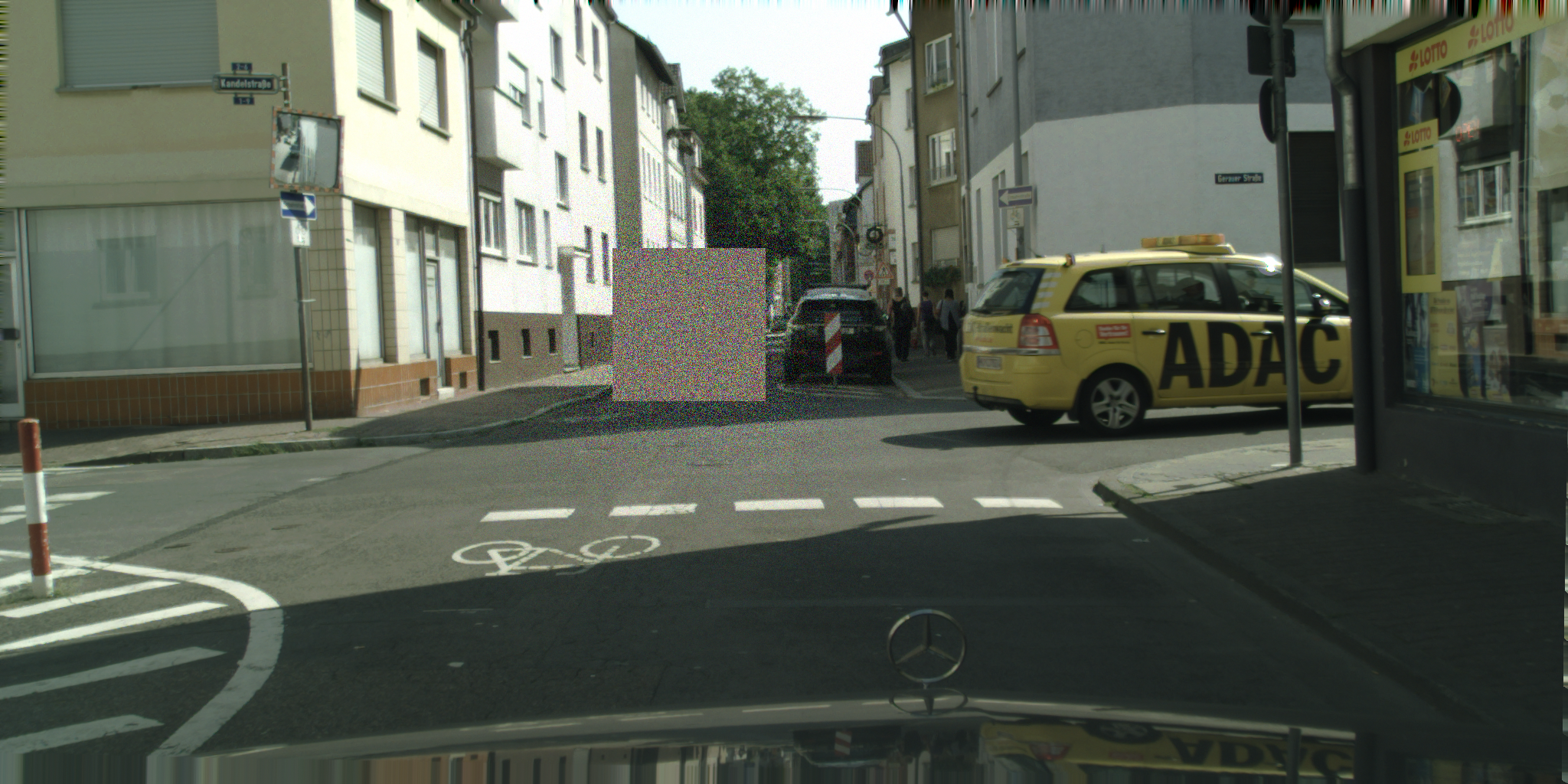}
    \subcaption{Patched Image}
\end{minipage} &
\begin{minipage}[h]{0.23\linewidth}
    \centering
    \includegraphics[width=\linewidth]{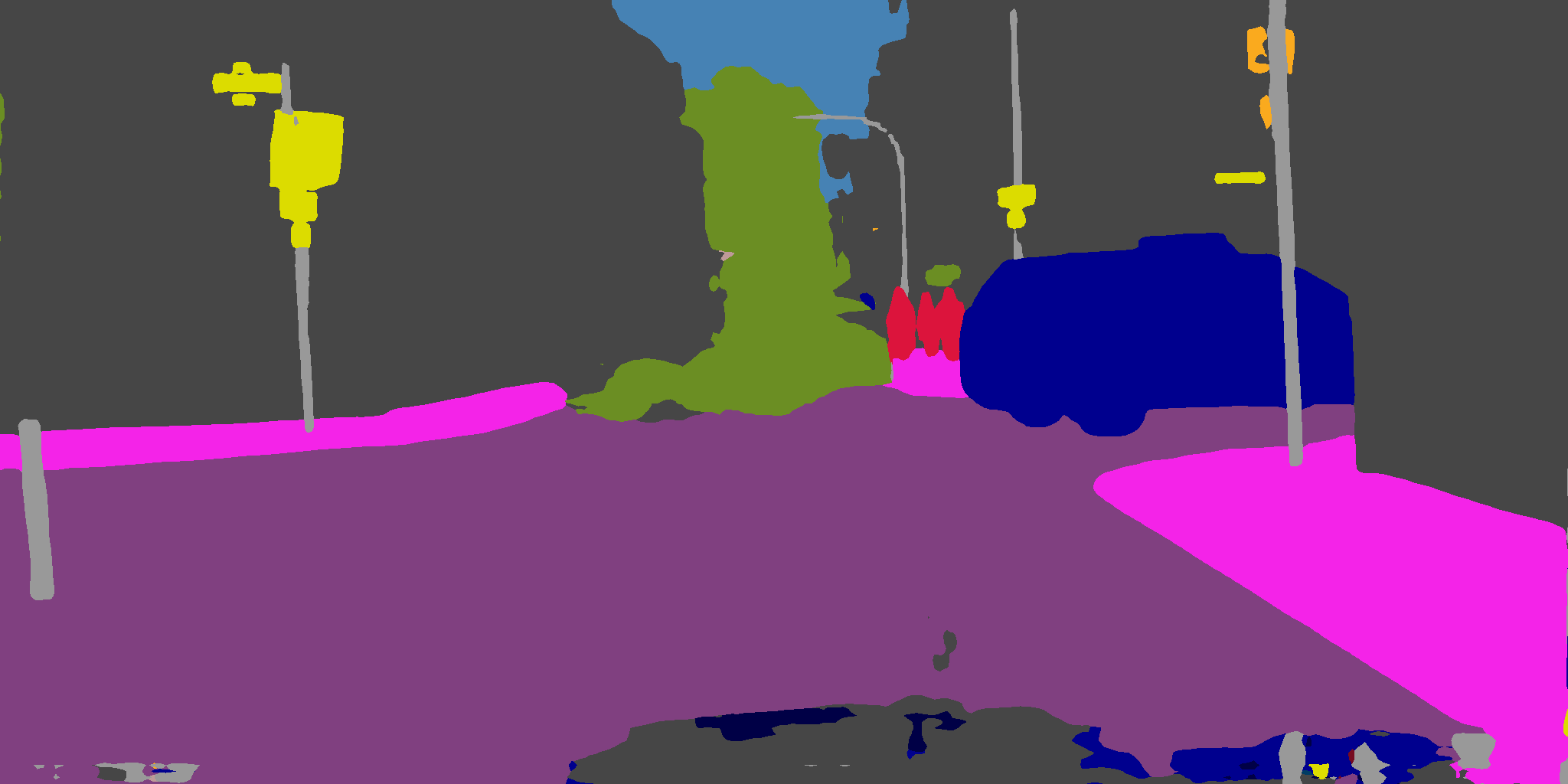}
    \subcaption{PIDNet-S}
\end{minipage} &
\begin{minipage}[h]{0.23\linewidth}
    \centering
    \includegraphics[width=\linewidth]{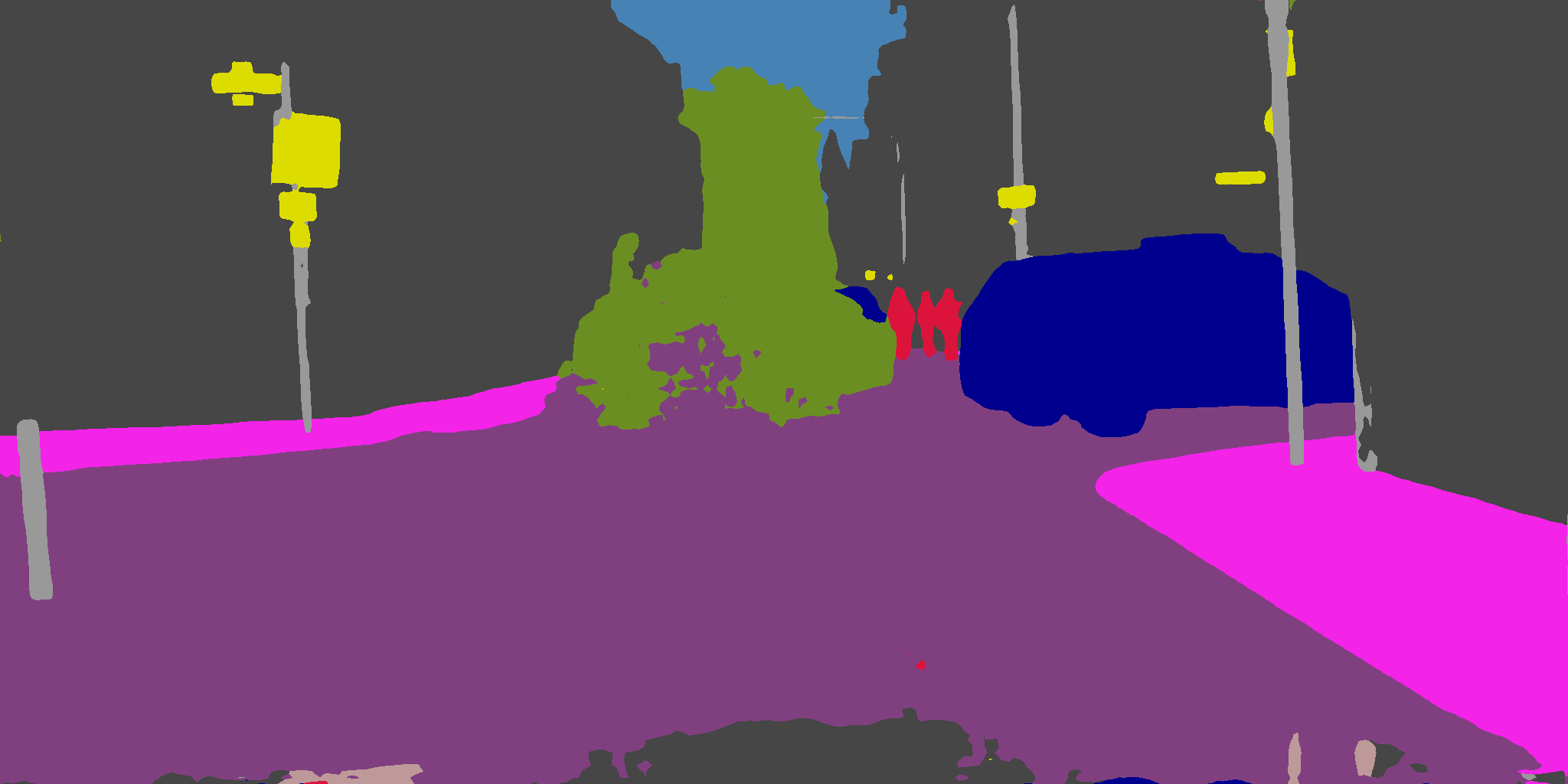}
    \subcaption{PIDNet-M}
\end{minipage} &
\begin{minipage}[h]{0.23\linewidth}
    \centering
    \includegraphics[width=\linewidth]{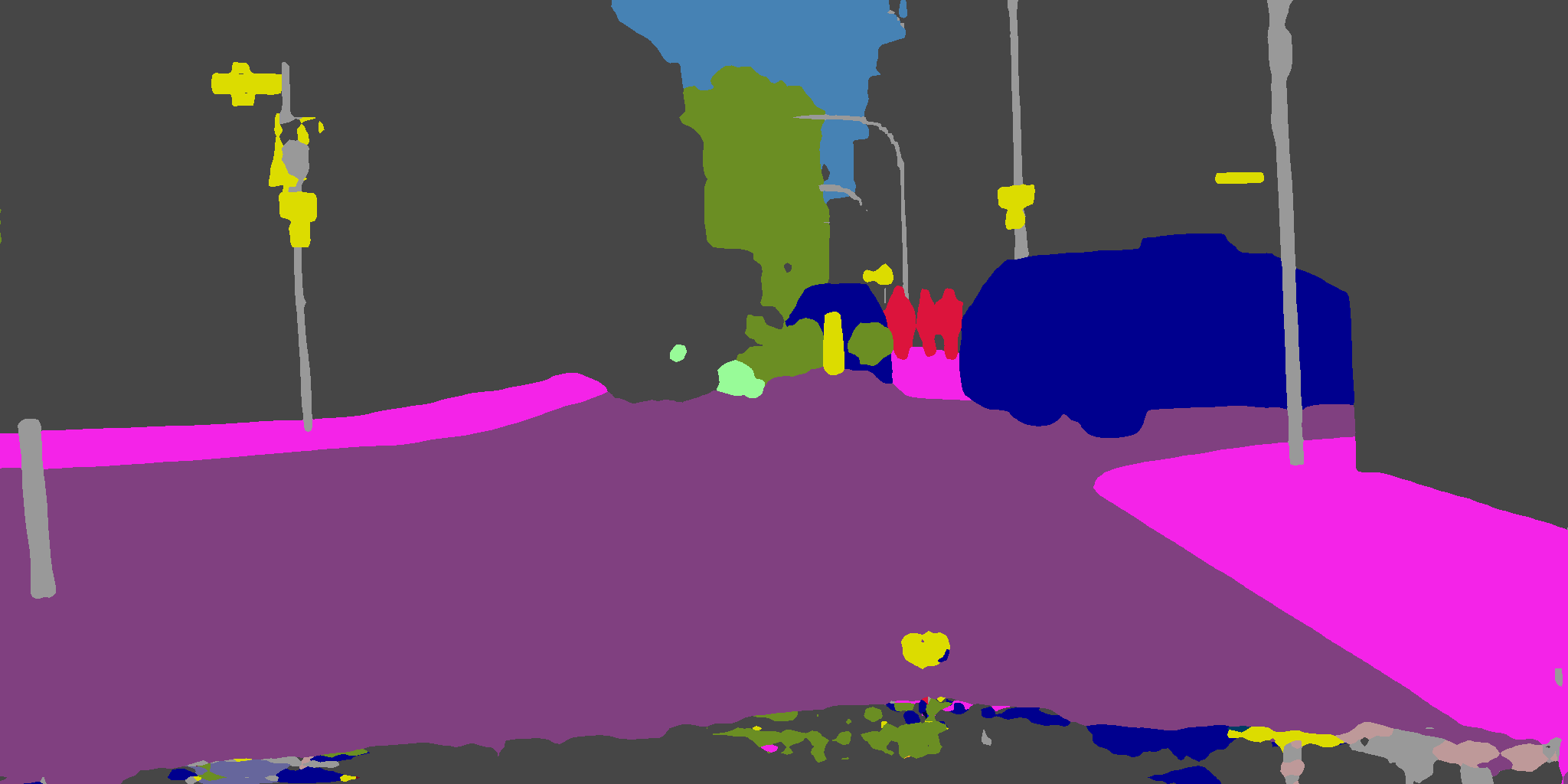}
    \subcaption{PIDNet-L}
\end{minipage} \\[-2pt]  % Negative spacing to pull rows together

% --------- ROW 2 ----------
\begin{minipage}[h]{0.23\linewidth}
    \centering
    \includegraphics[width=\linewidth]{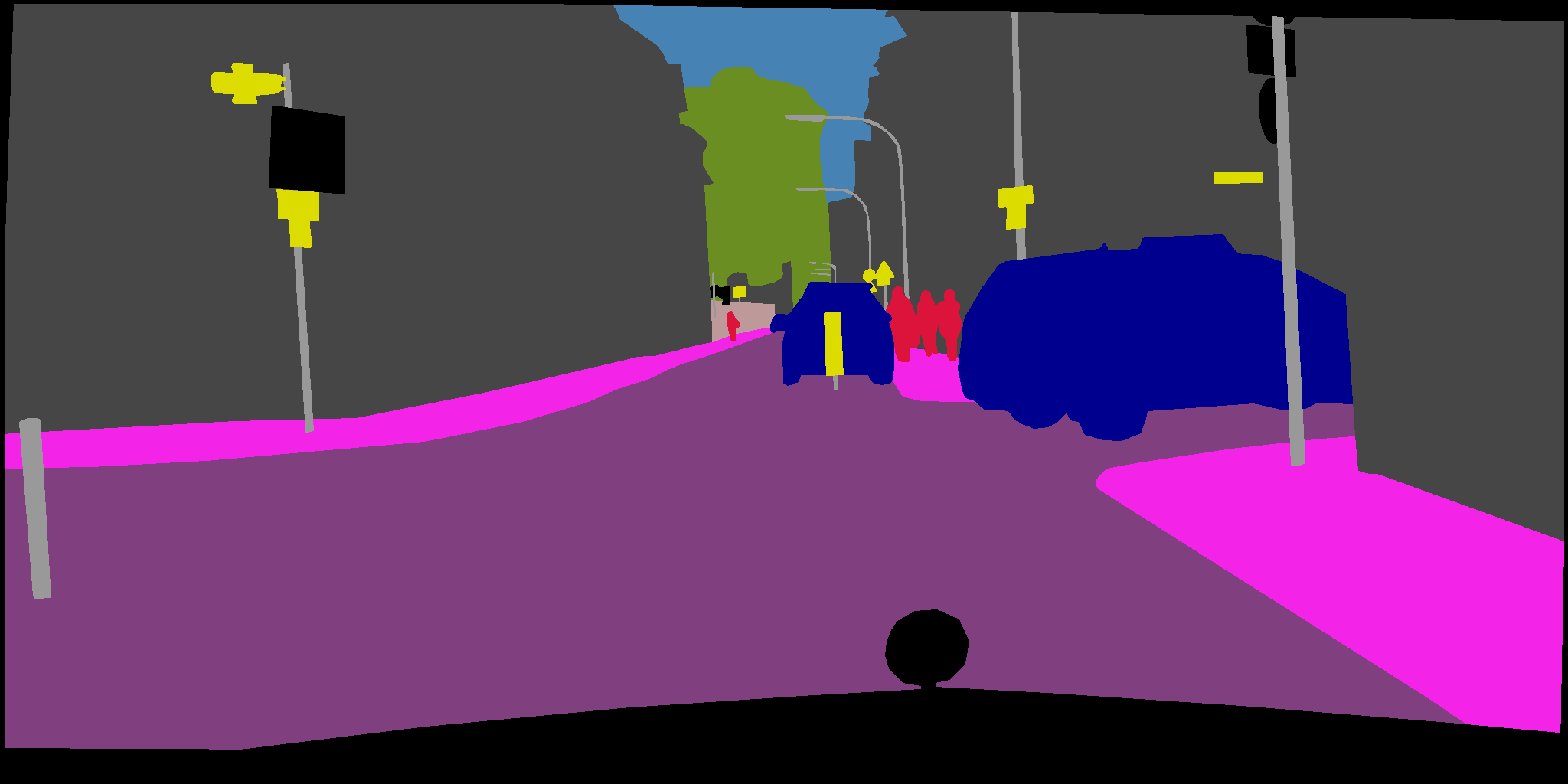}
    \subcaption{Ground Truth}
\end{minipage} &
\begin{minipage}[h]{0.23\linewidth}
    \centering
    \includegraphics[width=\linewidth]{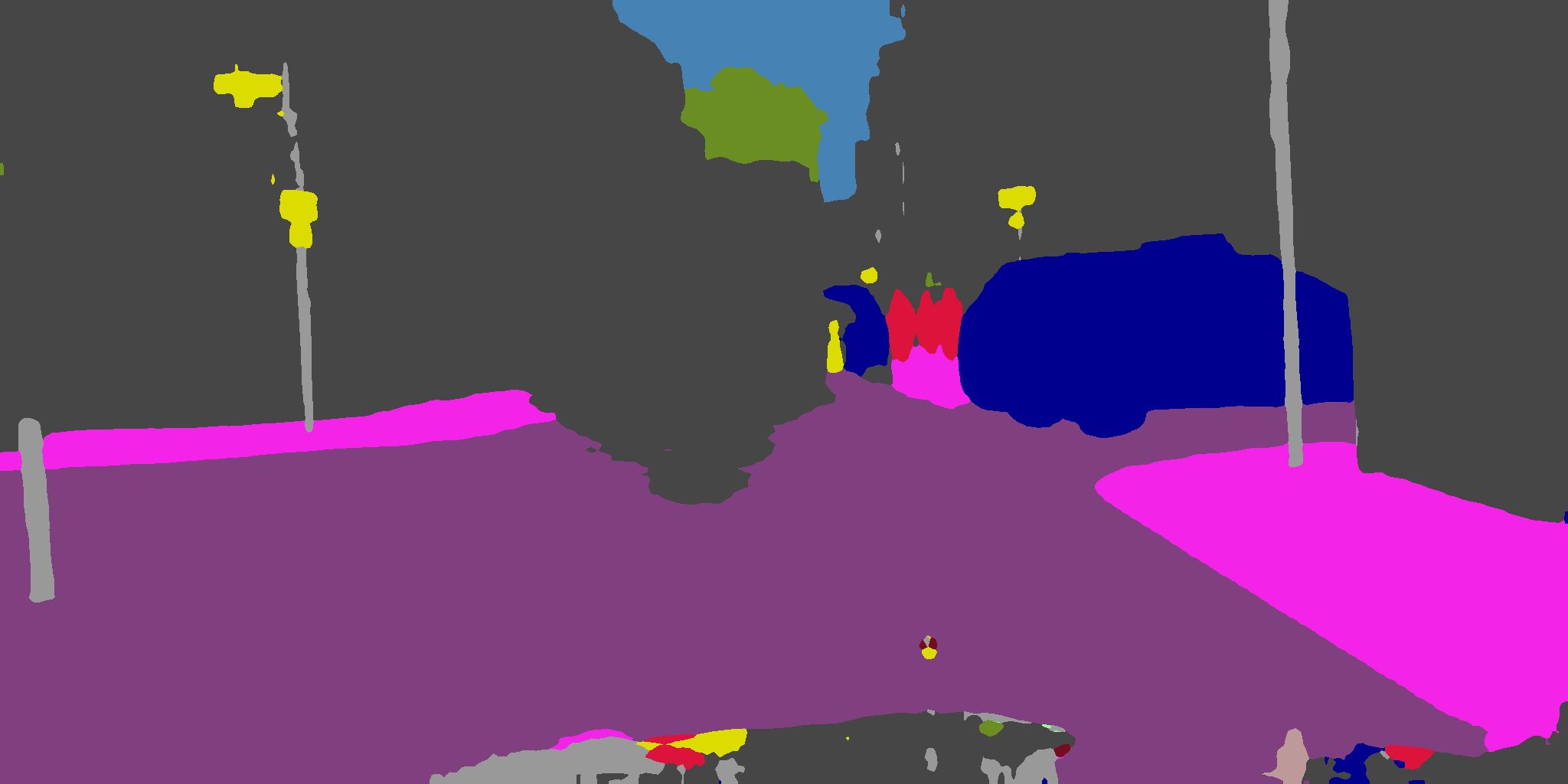}
    \subcaption{BiSeNetV1}
\end{minipage} &
\begin{minipage}[h]{0.23\linewidth}
    \centering
    \includegraphics[width=\linewidth]{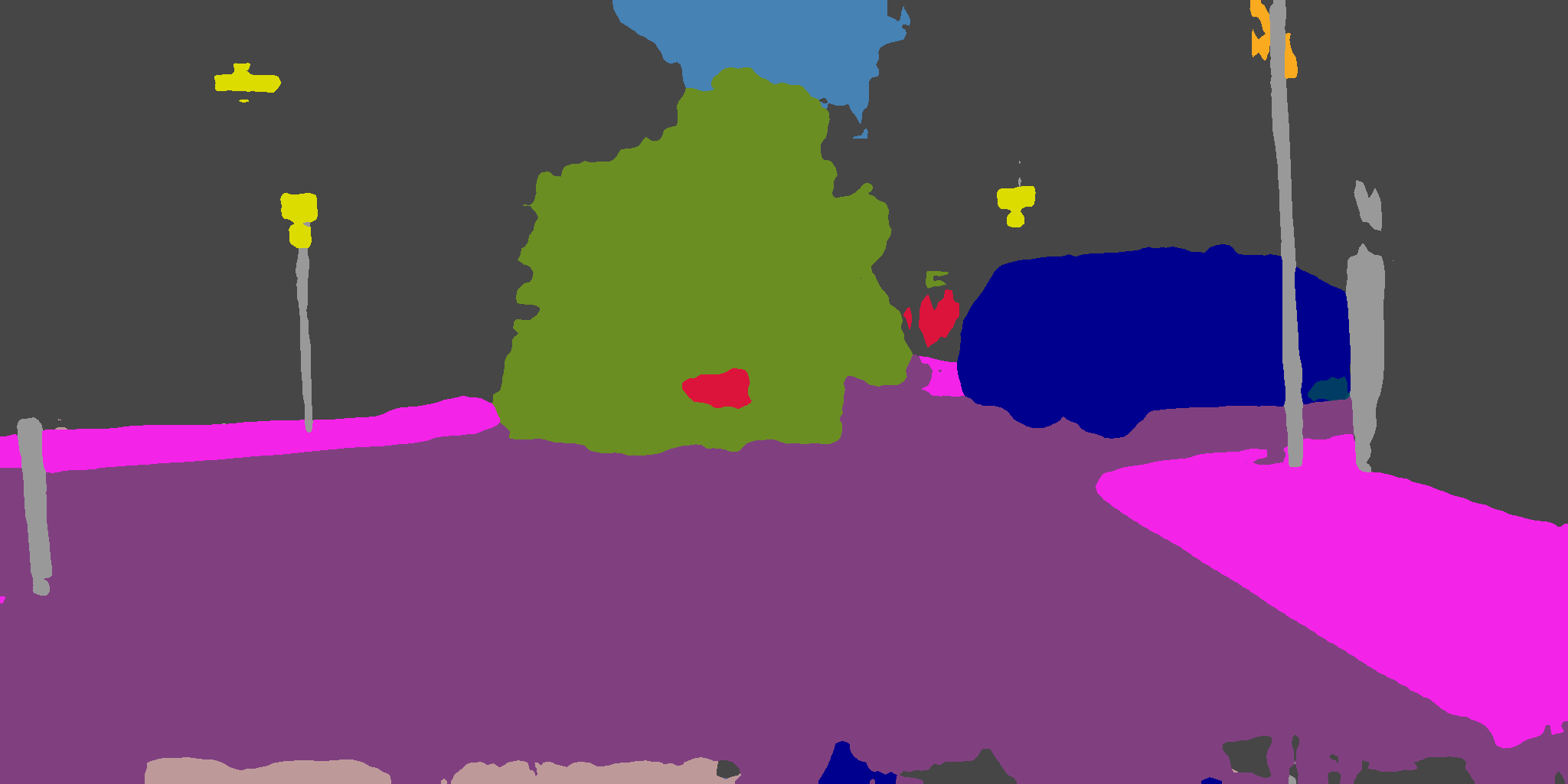}
    \subcaption{BiSeNetV2}
\end{minipage} &
\begin{minipage}[h]{0.23\linewidth}
    \centering
    \includegraphics[width=\linewidth]{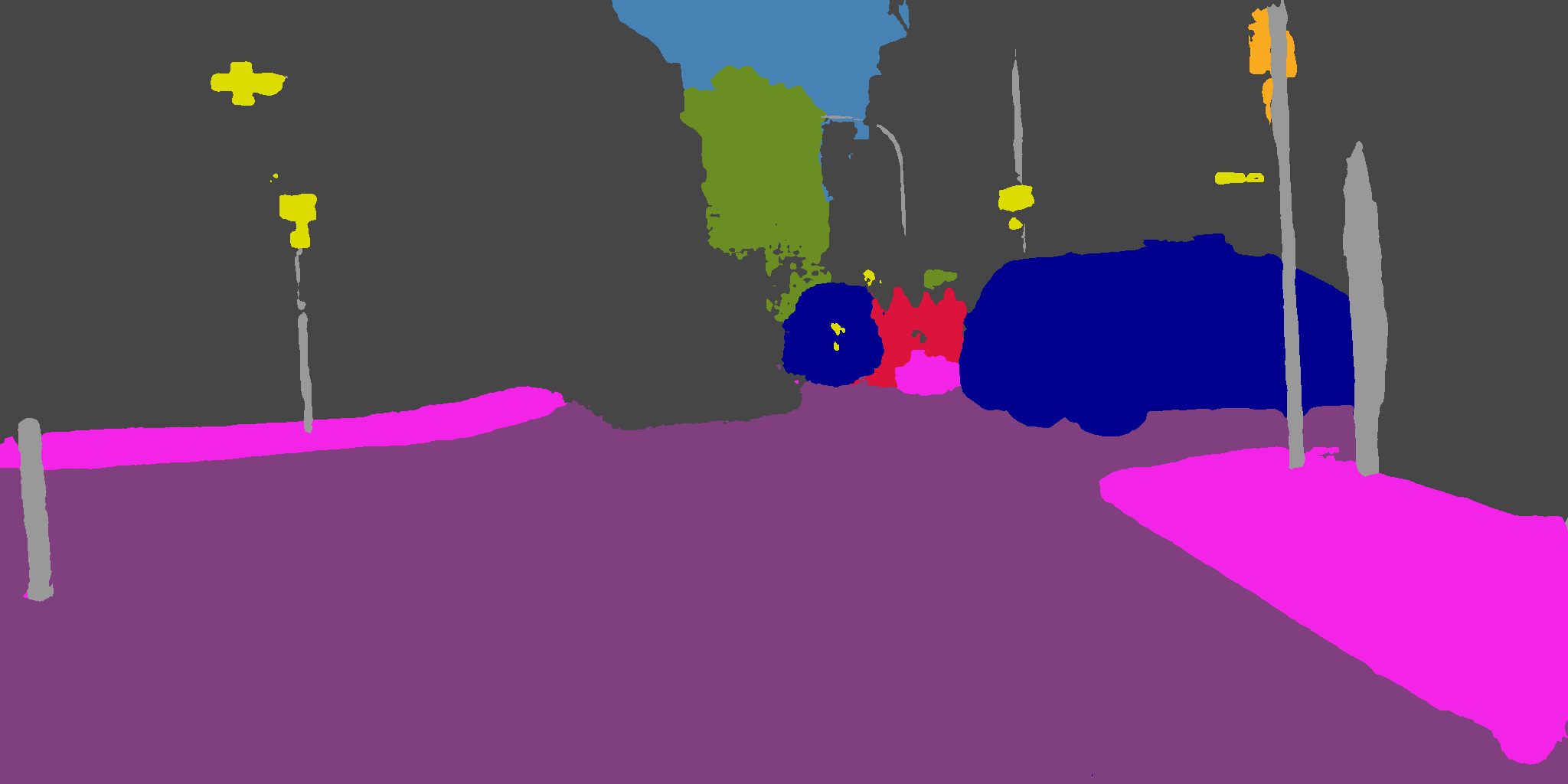}
    \subcaption{SegFormer}
\end{minipage}

\end{tabular}
\end{minipage}
\caption{Qualitative comparison of segmentation predictions across different models on the Cityscapes dataset. The perturbations are centered around the OmniPatch in all models.}
\label{fig:qualitative}
\end{figure*}

\subsection{Results}
We apply and train the patch on the `pole' label (most sensitive class in Cityscapes training data). We measure against the PIDNet-S trained patch in \citet{shekhar2025cross} which is used as baseline\footnote{Resource constraints required input downscaling for SegFormer, making baseline comparisons infeasible.}.
\begin{table}[htbp]
    \caption{mIoU on Cityscapes, comparing OmniPatch against clean, random and baseline.}
    \centering
    % Adds horizontal space between columns
    \setlength{\tabcolsep}{4pt} 
    % Adds vertical space between rows
    \renewcommand{\arraystretch}{0.8} 
    
    \begin{tabular}{|l|c|c|c|c|c|c|}
    \hline
    \textbf{Model} & 
    \shortstack{\textbf{Clean Image}\\\textbf{mIoU}} & 
    \shortstack{\textbf{Random}\\\textbf{Patch mIoU}} & 
    \shortstack{\textbf{Baseline}\\\textbf{Patch mIoU}} & 
    \shortstack{\textbf{OmniPatch}\\\textbf{mIoU}} & 
    \shortstack{\textbf{mIoU Drop (\%)}\\\textbf{ Clean / Random/Baseline}} \\ \hline
    
    PIDNet-S   & 0.8695 & 0.8651 & 0.7791 & \textbf{0.7299} & 16.05 / 15.62/ 6.31 \\ \hline
    PIDNet-M   & 0.8681 & 0.8618 & 0.8615 & \textbf{0.7393} & 14.84 / 14.22/ 14.18  \\ \hline
    PIDNet-L   & 0.9035 & 0.8996 & 0.8984 & \textbf{0.7530} & 16.65 / 16.30/ 16.18  \\ \hline
    BiSeNetv1  & 0.7149 & 0.7057 & 0.7120 & \textbf{0.6410} & 10.33 / 9.17/ 9.97  \\ \hline
    BiSeNetv2  & 0.6907 & 0.6845 & 0.6848 & \textbf{0.6036} & 12.61 / 11.81/ 11.85\\ \hline
    SegFormer  & 0.7434 & 0.7431& - & \textbf{0.6777} & 8.83 / 8.79/ - \\ \hline
    \end{tabular}
    \label{tab:miou_comparison}
\end{table}

% We also perform our experiments on the BDD100K dataset \citep{yu2020bdd100k}. 

% \begin{figure}
%     \centering
%     \includegraphics[width=1\linewidth]{segmentation_graphs.png}
%     \caption{Segmentation Maps}
%     \label{fig:placeholder}
% \end{figure}
% ========== REQUIRED IN PREAMBLE ==========
% \usepackage{graphicx}
% \usepackage{subcaption}
% \usepackage{multirow}
% \usepackage{adjustbox}  % Optional, for more control
% ==========================================

% ========== REQUIRED IN PREAMBLE ==========
% \usepackage{graphicx}
% \usepackage{subcaption}
% \usepackage{multirow}
% ==========================================

\subsection{Ablations}
To validate benefits of each step in our pipeline, we conduct ablation experiments (Figure \ref{fig:ablation_exp}) \footnote{{Detailed results are provided in Appendix~\ref{app:ablation_tables}}}.

\begin{figure}[hb]
\centering
\setlength{\tabcolsep}{3pt}
\captionsetup[subfigure]{skip=2pt, font=footnotesize}

\begin{tabular}{@{}ccc@{}}   % Removed extra column spacing

% --------- 3 IMAGES ----------
\begin{minipage}[hb]{0.26\linewidth}   % Increased from 0.28
    \centering
    \includegraphics[width=\linewidth]{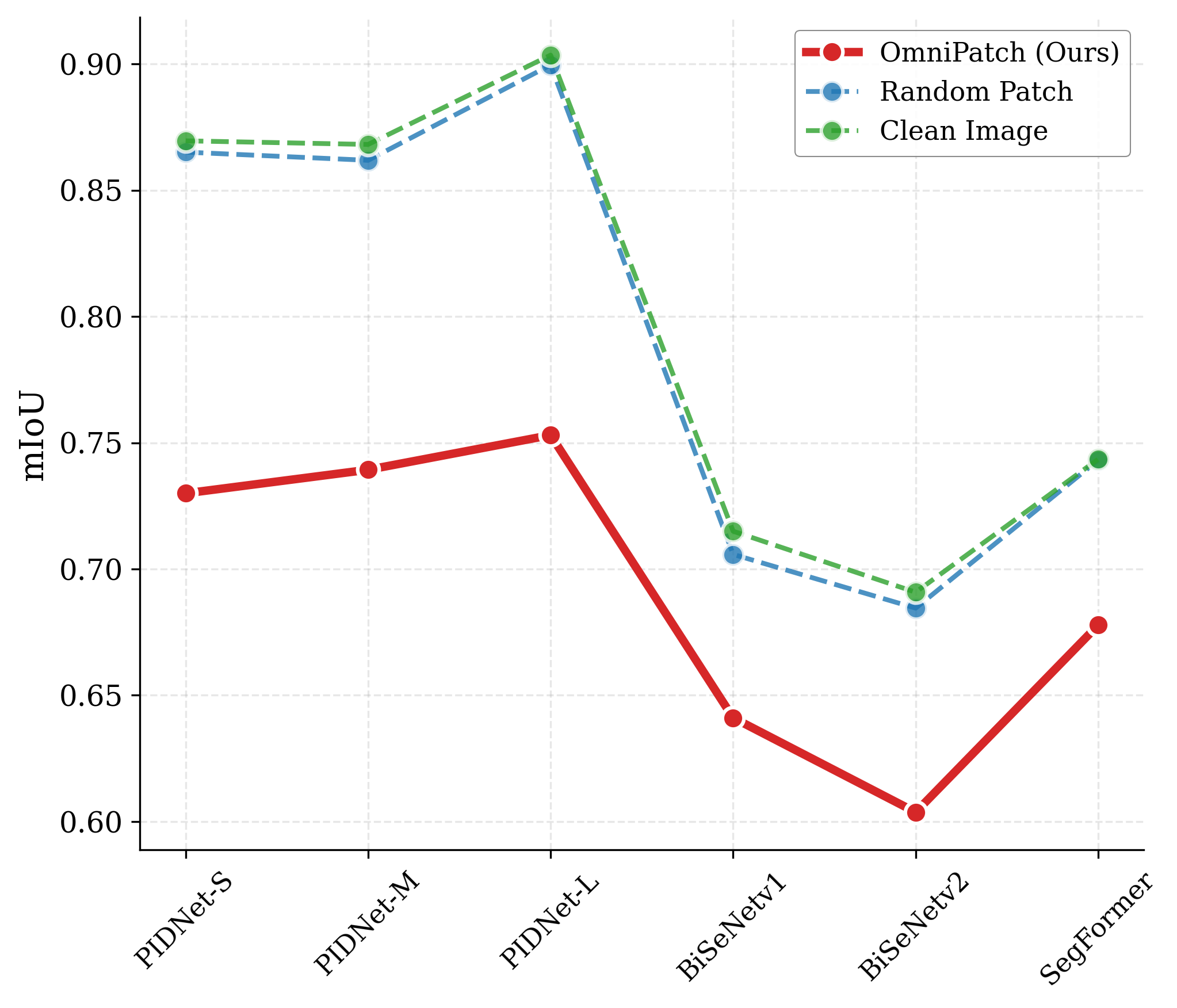}
    \subcaption{Across different patches}
\end{minipage} &
\begin{minipage}[hb]{0.26\linewidth}   % Increased from 0.28
    \centering
    \includegraphics[width=\linewidth]{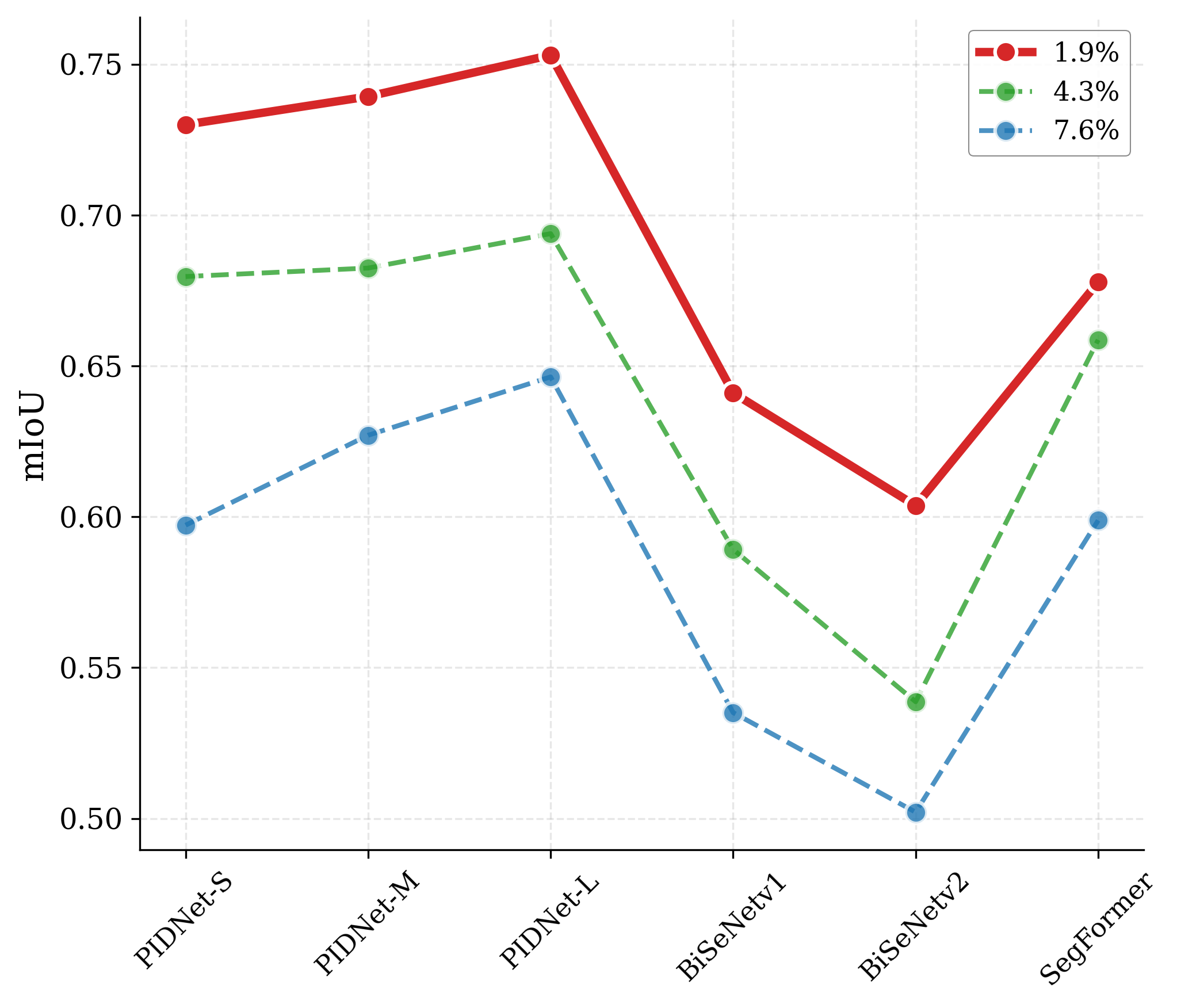}
    \subcaption{Patch area coverage}
\end{minipage} &
\begin{minipage}[hb]{0.26\linewidth}   % Increased from 0.28
    \centering
    \includegraphics[width=\linewidth]{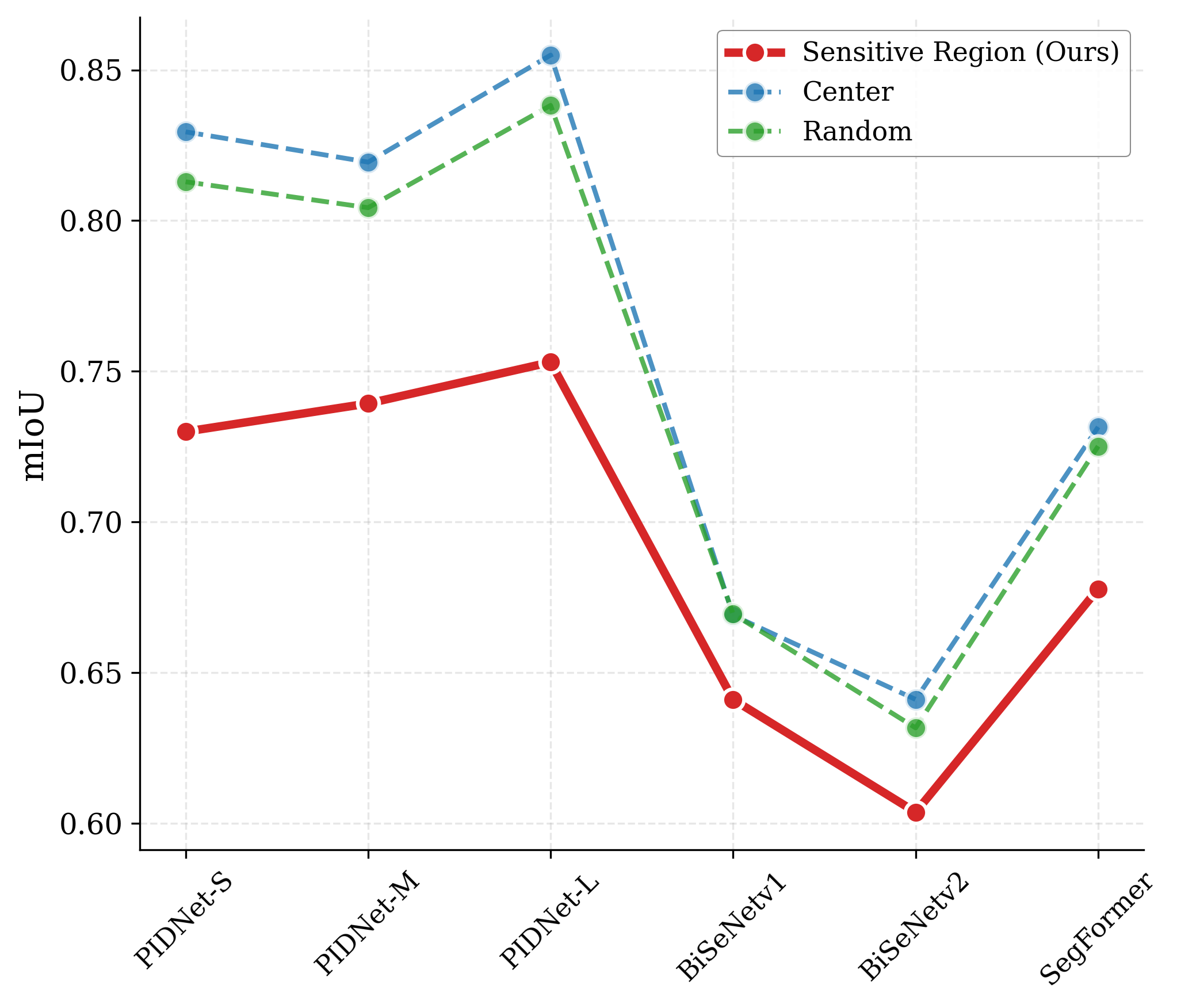}
    \subcaption{Patch Placement Strategy}
\end{minipage}
\end{tabular}
\caption{Component-wise ablation study analyzing the sensitivity of OmniPatch.}
\label{fig:ablation_exp}
\end{figure}
We observe that sensitive region placement over center and random improves performance. This highlights the importance of strategic spatial positioning. We note a positive correlation between patch size and mIoU drop, suggesting the increment in perturbations is due to larger coverage. We use JS divergence to measure distribution shift due to its boundedness, substituting KL with JS divergence enables stable training, yielding an additional average mIoU drop of \textbf{1.84\%} over models.
\section{Future Work and Limitations}
While OmniPatch bridges the gap between theoretical image-wide perturbations and formulates deployable attacks, it introduces visually obtrusive noise which is noticeable. Future research will explore concealment by developing texture blending techniques. We also aim to develop a patched attack functional in varying weather and lighting conditions instead of homogeneous images. Further, actual physical experiments should also be conducted for a conclusive proof-of-concept.
\section{Conclusion}
In this paper, we focus on a model-agnostic and conceptually deployable adversarial attack on semantic segmentation models. We successfully formulate a pipeline to design such an attack and prove its effectiveness on both ViT and CNN based segmentation models. We introduce a novel uncertainty-based spatial positioning scheme. We extend the work of \citet{jia2023transegpgd} by introducing a CNN surrogate to employ ensemble learning for increased transferability. We also include subsidiary regularizers and perform comprehensive experiments and ablations on diverse models.

\bibliography{iclr2026_conference}
\bibliographystyle{iclr2026_conference}

\newpage
\appendix

\section{Sensitive Region Placement}
\label{app:method_1}

\subsection{Sensitive Class Identification}
\label{app:method_1:sensitive_region}

To identify the class which is most susceptible to adversarial perturbations, we utilize predictive entropy as a proxy for model uncertainty. High-entropy regions indicate areas where the model's decision boundary is fragile, making them ideal targets for attack.

We extract the pixel-wise probability distributions from the ViT surrogate. Let $p_{b,c,h,w}$ denote the predicted probability for class $c$ at pixel $(h,w)$ in image $b$. We first compute the normalized pixel-wise entropy map $H_{b,h,w}$:
\begin{equation}
H_{b,h,w} = - \frac{1}{\log C} \sum_{c=1}^{C} p_{b,c,h,w} \log \left( p_{b,c,h,w} + \epsilon \right).
\end{equation}

Next, we aggregate this uncertainty for each class within the image. Let $\mathbf{1}_{\{y_{b,h,w} = c\}}$ be an indicator function that is 1 if pixel $(h,w)$ belongs to class $c$. The mean entropy for class $c$ in image $b$ is computed as:
\begin{equation}
\bar{H}_{b,c} =
\frac{\sum_{h,w} \mathbf{1}_{\{y_{b,h,w} = c\}} H_{b,h,w}}
{\sum_{h,w} \mathbf{1}_{\{y_{b,h,w} = c\}} + \epsilon}.
\end{equation}
Finally, we compute the global sensitivity score $S_c$ by averaging $\bar{H}_{b,c}$ over the entire dataset of $B$ images. The target class $c^\star$ is selected as the one maximizing this global uncertainty:
\begin{equation}
S_{c} = \frac{1}{B} \sum_{b=1}^{B} \bar{H}_{b,c}, \quad c^\star = \arg \max_{c} S_c.
\end{equation}

Based on our experimental configuration, we observed the following per-class sensitivity scores
\begin{figure}[h]
    \centering
    \includegraphics[width=1\linewidth]{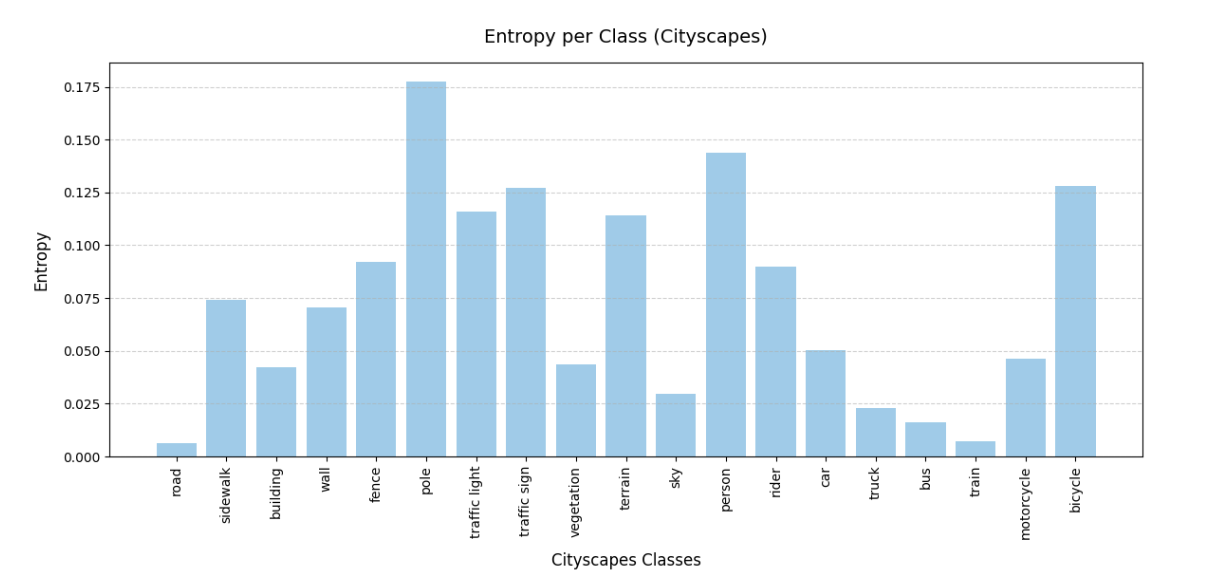}
    \caption{Class-wise Sensitivity Score across Cityscapes Dataset}
    \label{fig:placeholder}
\end{figure}

\subsection{Morphological Dilation}
\label{app:method_1:morphological_dialation}

To expand the feasible region for patch placement we expand the binary mask $M \in \{0, 1\}^{H \times W}$ using morphological dilation.

Let $B_k$ be a square structuring element of size $k \times k$. The dilated mask $\tilde{M}$ is defined as the Minkowski sum of the original mask $M$ and the structuring element $B_k$:
\begin{equation}
\tilde{M} = M \oplus B_k = \left\{ z \in \Omega \mid (B_k)_z \cap M \neq \emptyset \right\},
\end{equation}
where $(B_k)_z$ denotes the translation of $B_k$ by vector $z$. In our implementation, we perform this operation efficiently using a max-pooling layer with kernel size $k$ and stride 1:
\begin{equation}
\tilde{M}_{h,w} = \max_{(i,j) \in \mathcal{N}_k(h,w)} M_{i,j},
\end{equation}
where $\mathcal{N}_k(h,w)$ represents the $k \times k$ spatial neighborhood centered at $(h,w)$. 
% We use $k=5$ to capture the immediate context surrounding the object.

\subsection{Biased Sampling for Patch Placement}
\label{app:method_1:biased_sampling}
To automate the selection of optimal patch locations, we implement an entropy-biased sampling strategy. Let $\tilde{M} \in \{0, 1\}^{H \times W}$ be the dilated binary mask of the target class and $H(\cdot)$ be the pixel-wise entropy map. The patch top-left coordinate $(y_0, x_0)$ is determined via the following procedure:
\paragraph{1. Feasible Region Definition.}
We first define the set of valid center coordinates $\mathcal{V}$ to ensure the patch of size $S \times S$ remains fully within the image boundaries. A pixel $(y, x)$ is considered valid if it lies within the mask $\tilde{M}$ and satisfies the boundary constraints:
\begin{equation}
\mathcal{V} = \left\{ (y, x) \mid \tilde{M}_{y,x} = 1, \quad \frac{S}{2} \le y \le H - \frac{S}{2}, \quad \frac{S}{2} \le x \le W - \frac{S}{2} \right\}.
\end{equation}

\paragraph{2. Entropy Filtering.}
We extract the entropy values for all valid pixels: $E = \{ H(y, x) \mid (y, x) \in \mathcal{V} \}$. Let $\tau$ be the $(1-p)^{\text{th}}$ quantile of $E$, where $p$ represents the sampling fraction. The high-uncertainty candidate set $\mathcal{V}_{\text{top}}$ is defined as:
\begin{equation}
\mathcal{V}_{\text{top}} = \left\{ (y, x) \in \mathcal{V} \mid H(y, x) \ge \tau \right\}.
\end{equation}

\paragraph{3. Uniform Sampling.}
We sample a center coordinate $(y_c, x_c)$ from $\mathcal{V}_{\text{top}}$ uniformly. The probability of choosing any pixel $(y, x)$ as the center is given by:
\begin{equation}
P(y, x) =
\begin{cases}
\frac{1}{|\mathcal{V}_{\text{top}}|}, & \text{if } (y, x) \in \mathcal{V}_{\text{top}} \\
0, & \text{otherwise}
\end{cases}
\end{equation}
The final top-left coordinate is derived as $(y_0, x_0) = (y_c - \lfloor S/2 \rfloor, x_c - \lfloor S/2 \rfloor)$.

\section{Detailed Experimental Details}

\subsection{Detailed Results}
\label{app:ablation_tables}
This section presents comprehensive experimental results and additional quantitative analyses that were omitted from the main text due to space constraints. We provide a detailed breakdown of the component-wise ablation study, validating the specific contribution of each loss term.

\begin{table}[htbp]
    \caption{Performance Comparison across Different Patch Attacks}
    \centering
    % Adds horizontal space between columns
    \setlength{\tabcolsep}{8pt} 
    % Adds vertical space between rows
    \renewcommand{\arraystretch}{1.2} 
    \begin{tabular}{lccc}
        \hline
        \textbf{Model} & \textbf{Clean Image} & \textbf{Random Patch} & \textbf{OmniPatch} \\
        \hline
        PIDNet-S & 0.869577 & 0.865101 & 0.729970 \\
        PIDNet-M & 0.868144 & 0.861862 & 0.739300 \\
        PIDNet-L & 0.903522 & 0.899625 & 0.753016 \\
        BiSeNetv1 & 0.714959 & 0.705789 & 0.641062 \\
        BiSeNetv2 & 0.690796 & 0.684500 & 0.603687 \\
        SegFormer & 0.743495 & 0.743120 & 0.677788 \\
        \hline
    \end{tabular}
    \label{tab:model_performance}
\end{table}
\begin{table}[htbp]
    \caption{Model Performance under varying Patch areas}
    \centering
    % Adds horizontal space between columns
    \setlength{\tabcolsep}{8pt} 
    % Adds vertical space between rows
    \renewcommand{\arraystretch}{1.2} 
    \begin{tabular}{lccc}
        \hline
        \textbf{Model} & \textbf{Patch 200 (1.9\%)} & \textbf{Patch 300 (4.3\%)} & \textbf{Patch 300 (7.6\%)} \\
        \hline
        PIDNet-S  & 0.729970 & 0.679634 & 0.597220 \\
        PIDNet-M  & 0.739300 & 0.682442 & 0.627030 \\
        PIDNet-L  & 0.753016 & 0.693860 & 0.646377 \\
        BiSeNetv1 & 0.641062 & 0.589184 & 0.535108 \\
        BiSeNetv2 & 0.603687 & 0.538712 & 0.502113 \\
        SegFormer & 0.677788 & 0.658646 & 0.598823 \\
        \hline
    \end{tabular}
    \label{tab:patch_size_comparison}
\end{table}
\begin{table}[htbp]
    \caption{Impact of Patch Placement Strategies on Model Robustness}
    \centering
    % Adds horizontal space between columns
    \setlength{\tabcolsep}{8pt} 
    % Adds vertical space between rows
    \renewcommand{\arraystretch}{1.2} 
    \begin{tabular}{lccc}
        \hline
        \textbf{Model} & \textbf{Center Placement} & \textbf{Random Placement} & \textbf{Sensitive Region (Ours)} \\
        \hline
        PIDNet-S  & 0.829445 & 0.812880 & 0.729970 \\
        PIDNet-M  & 0.819404 & 0.804276 & 0.739300 \\
        PIDNet-L  & 0.854908 & 0.838226 & 0.753016 \\
        BiSeNetv1 & 0.668978 & 0.669475 & 0.641062 \\
        BiSeNetv2 & 0.641095 & 0.631650 & 0.603687 \\
        SegFormer & 0.731485 & 0.725132 & 0.677788 \\
        \hline
    \end{tabular}
    \label{tab:placement_comparison}
\end{table}
\begin{table}[htbp]
    \caption{Comparative Impact of KL and JS Divergence on Adversarial Patch Effectiveness}
    \centering
    % Adds horizontal space between columns
    \setlength{\tabcolsep}{8pt} 
    % Adds vertical space between rows
    \renewcommand{\arraystretch}{1.2} 
    \begin{tabular}{lcc}
        \hline
        \textbf{Model} & \textbf{With KL divergence} & \textbf{With JS divergence (Ours)} \\
        \hline
        PIDNet-S  & 0.746490 & 0.729970 \\
        PIDNet-M  & 0.746948 & 0.739300 \\
        PIDNet-L  & 0.766492 & 0.753016 \\
        BiSeNetv1 & 0.646691 & 0.641062 \\
        BiSeNetv2 & 0.606668 & 0.603687 \\
        SegFormer & 0.708983 & 0.677788 \\
        \hline
    \end{tabular}
    \label{tab:divergence_comparison}
\end{table}
\begin{table}[htbp]
    \caption{Impact of gradient alignment on adversarial patch effecitveness (mIoU)}
    \centering
    % Adds horizontal space between columns
    \setlength{\tabcolsep}{8pt} 
    % Adds vertical space between rows
    \renewcommand{\arraystretch}{1.2} 
    \begin{tabular}{lcc}
        \hline
        \textbf{Model} & \textbf{Without Gradient Alignment} & \textbf{With Gradient Alginment (Ours)} \\
        \hline
        PIDNet-S &   0.777234 & 0.729970    \\
        PIDNet-M &   0.765665 & 0.739300  \\
        PIDNet-L &   0.786792  & 0.753016  \\
        BiSeNetv1&  0.662936   & 0.641062 \\
        BiSeNetv2&  0.627358 & 0.603687    \\
        SegFormer&  0.680991 & 0.677788\\
        \hline
    \end{tabular}
    \label{tab:divergence_comparison}
\end{table}

% \begin{table}[htbp]
%     \caption{Impact of Different Auxiliary losses}
%     \centering
%     % Adds horizontal space between columns
%     \setlength{\tabcolsep}{8pt} 
%     % Adds vertical space between rows
%     \renewcommand{\arraystretch}{1.2} 
%     \begin{tabular}{lcccc}
%         \hline
%         \textbf{Model} & \textbf{Full} & \textbf{w/o attention} & \textbf{w/o boundary} & \textbf{w/o total} \\
%         & \textbf{Method} & \textbf{hijack} & \textbf{disruption} & \textbf{variation} \\
%         \hline
%         PIDNet-S  & 0.729970 & 0.814804 & 0.814803 & 0.814804 \\
%         PIDNet-M  & 0.739300 & 0.812684 & 0.812683 & 0.812683 \\
%         PIDNet-L  & 0.753016 & 0.844478 & 0.844479 & 0.844479 \\
%         BiSeNetv1 & 0.641062 & 0.669861 & 0.669854 & 0.669798 \\
%         BiSeNetv2 & 0.603687 & 0.652981 & 0.652980 & 0.652981 \\
%         SegFormer & 0.677788 & 0.722656 & 0.722656 & 0.722656 \\
%         \hline
%     \end{tabular}
%     \label{tab:ablation_components}
% \end{table}

\newpage
\subsection{Reproducibility}
This section contains all necessary details which allow reproduction of results mentioned in the paper.

All experiments reported in this work are fully reproducible. The complete implementation, along with the exact hyperparameter settings used in our experiments, is provided in the accompanying \href{https://github.com/dsgiitr/omnipatch}{\textcolor{blue}{Code Repository}}.

\begin{table}[htbp]
    \caption{Implementation Details}
    \centering
    % Adds horizontal space between columns
    \setlength{\tabcolsep}{8pt}
    % Adds vertical space between rows
    \renewcommand{\arraystretch}{1.2}
    \begin{tabular}{lc}
        \hline
        \textbf{Parameter} & \textbf{Value} \\
        \hline

        \multicolumn{2}{c}{\textbf{Senstive Region Placement}} \\
        Morphological Dilation pixels $(k)$ & $5$\\
        Biased Sampling threshold $(p)$ & $0.2$\\
        \hline

        \multicolumn{2}{c}{\textbf{Stage 1}} \\
        Stage 1 Epochs & $10$ \\
        Stage 1 weighting $(\gamma)$ & $0.7$\\
        \hline

        \multicolumn{2}{c}{\textbf{Stage 2}} \\
        Stage 2 Epochs & $10$ \\
        Stage 2 weighting $(\beta)$ & $0.3$ \\
        JS Divergence threshold & $>\mu$ (Mean) \\
        Gradient Alignment (\texttt{grad\_align\_w}) & $1.0 \times 10^{-1}$ \\
        \hline
        \multicolumn{2}{c}{\textbf{Auxillary Objectives (Coefficients)}} \\
        Total Variation (\texttt{tv\_weight})       & $1.0 \times 10^{-4}$ \\
        Attention Hijack (\texttt{attn\_hijack\_w}) & $1.0 \times 10^{-1}$ \\
        Boundary (\texttt{boundary\_w})             & $2.0 \times 10^{-1}$ \\
        \hline
        \multicolumn{2}{c}{\textbf{Miscellaneous}} \\
        Batches per epoch & $150$\\
        Batch Size & $2$ \\
        Image Resolution & $2048 \times 1024$ \\
        SegFormer Downscaling & $0.75$ \\
        \hline
    \end{tabular}
    \label{tab:misc_details}
\end{table}

\end{document}